%% file: main.tex
\documentclass[journal]{IEEEtran}
\IEEEoverridecommandlockouts

\usepackage{amsmath,amsfonts}
\usepackage{algorithmic}
\usepackage{algorithm}
\usepackage{array}
\usepackage{multirow}
\usepackage{multicol}
\usepackage[caption=false,font=normalsize,labelfont=sf,textfont=sf]{subfig}
\usepackage{textcomp}
\usepackage{subcaption}
\usepackage{stfloats}
\usepackage{url}
\usepackage{verbatim}
\usepackage{graphicx}
\usepackage{cite}
\usepackage{hyperref}
\usepackage{xcolor}
\usepackage{placeins}

\newcommand{\vect}[1]{\mathbf{#1}}   

\DeclareUnicodeCharacter{2212}{\textendash}

\begin{document}

\title{Bridging Simplicity and Sophistication using GLinear: A Novel Architecture for Enhanced Time Series Prediction}

\author{Syed Tahir Hussain Rizvi$^1$\href{https://orcid.org/0000-0002-2656-6470}{\includegraphics[scale=0.01]{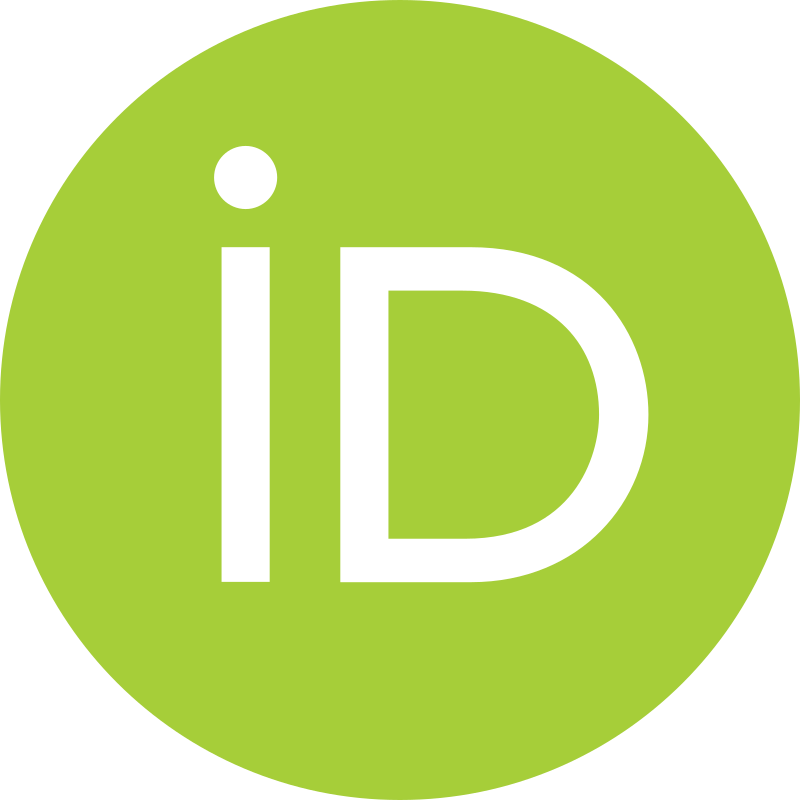}}, Neel Kanwal$^{1*}$\href{https://orcid.org/0000-0002-8115-0558}{\includegraphics[scale=0.01]{orcid.png}}, Muddasar Naeem$^2$\href{https://orcid.org/0000-0003-0815-4883}{\includegraphics[scale=0.01]{orcid.png}}\\
$^1$ Department of Electrical Engineering and Computer Science, University of Stavanger, 4021 Stavanger, Norway\\
$^2$ Research Center on ICT Technologies for Healthcare and Wellbeing, Università Telematica Giustino Fortunato, 82100 Benevento, Italy\\
{* Corresponding author: neel.kanwal@uis.no}}

\maketitle
\begin{abstract}
Time Series Forecasting (TSF) is an important application across many fields. There is a debate about whether Transformers, despite being good at understanding long sequences, struggle with preserving temporal relationships in time series data. Recent research suggests that simpler linear models might outperform or at least provide competitive performance compared to complex Transformer-based models for TSF tasks. In this paper, we propose a novel data-efficient architecture, \textit{Gaussian-activated Linear model (GLinear)}, for multivariate TSF that exploits periodic patterns to provide better accuracy. It achieves higher prediction accuracy while requiring less historical data than other state-of-the-art linear predictors. Four different datasets (ETTh1, Electricity, Traffic, and Weather) are used to evaluate the performance of the proposed predictor. A performance comparison with state-of-the-art linear architectures (such as NLinear, DLinear, and RLinear) and transformer-based time series predictors (Autoformer) shows that the GLinear, despite being data efficient, outperforms the existing architectures in most cases of multivariate TSF while being competitive in others. We hope that the proposed GLinear model opens new fronts of research and development of simpler and more sophisticated architectures for data and computationally efficient time-series analysis. The source code is publicly available on \href{https://github.com/t-rizvi/GLinear/}{GitHub}. 
\end{abstract}



\section{Introduction}
Accurate forecasting has become increasingly valuable in today's data-driven world, where computational intelligence plays a key role in automated decision-making~\cite{buansing2020information, ouyang2019modeling}. Time Series Forecasting (TSF) tasks have various applications that impact diverse fields such as finance, healthcare, supply chain management, and climate science~\cite{li2024deep}. The ability to predict future trends based on historical data not only enhances decision-making processes but also drives innovation and operational efficiency. 

Forecasting is typically categorized into short-term, medium-term, and long-term predictions, each serving distinct purposes and employing tailored methodologies~\cite{wazirali2023state}. Short-term forecasting focuses on immediate needs; medium-term forecasting assists in strategic planning, while long-term forecasting aids in vision-setting and resource allocation~\cite{makridakis2018statistical}. Historically, traditional statistical methods like exponential smoothing~\cite{makridakis2018statistical} and ARIMA~\cite{ariyo2014stock} have dominated the forecasting arena for short-term TSF. Traditional methods capture trend and seasonality components in the data and perform well due to their simplicity and responsiveness to recent data~\cite{abraham2009short, ouyang2019modeling}. However, they struggle with medium- and long-term predictions, and this decline can be attributed to assumptions of linearity and stationarity, often leading to overfitting and limited adaptability~\cite{zhou2022hybrid, lin2021temporal}. 

The emergence of advanced and hybrid methods that use Deep Learning (DL) has revolutionized predictive modeling by enabling the extraction of complex patterns within the data \cite{Qayyumdiag24}. Recurrent Neural Networks (RNNs), Long Short-Term Memory (LSTM), and Transformer-based architectures have shown promising results in medium- and long-term forecasting without rigid assumptions~\cite{dai2025salstm, zeng2023transformers, tian2024psrunet}. Despite their sophistication, recent studies indicate that linear models can capture periodic patterns and provide competitive performance in certain contexts, along with computational efficiency~\cite{zeng2023transformers,li2023revisiting}.

Not all time-series data are suitable for precise predictions, particularly when it comes to long-term forecasting, which becomes especially difficult in chaotic systems~\cite{li2023revisiting}. Long-term forecasting is most feasible for time series data when data exhibits clear trends and periodic patterns ~\cite{li2023revisiting,zeng2023transformers}. This brings up a question: \textbf{\textit{How can we effectively integrate the simplicity of linear models with sophisticated techniques for capturing complex underlying patterns to further enhance medium- and long-term TSF?} }


Among popular linear models, \textit{NLinear}~\cite{zeng2023transformers} often struggles with non-linear relationships in data, leading to suboptimal performance in complex forecasting scenarios. On the other hand, \textit{DLinear}~\cite{zeng2023transformers} is computationally intensive and may require large amounts of training data, which can hinder real-time application and scalability. Although \textit{RLinear}~\cite{li2023revisiting} models are capable of capturing trends and seasonality, they often fall short in their ability to generalize across varying datasets.

Inspired by long TSF linear models (NLinear~\cite{zeng2023transformers}, DLinear~\cite{zeng2023transformers}, and RLinear~\cite{li2023revisiting}), we propose a novel data-efficient architecture, \textbf{\textit{Gaussian-activated Linear (GLinear) model}}. GLinear is a simple model that does not have any complex components, functions, or blocks (like self-attention schemes, positional encoding blocks, etc.), like previously mentioned linear models. It has capabilities of enhanced forecasting performance while maintaining simplicity. Furthermore, GLinear focuses on data efficiency and demonstrates the potential to perform robust forecasting without relying on extensive historical datasets, which is a common limitation of other TSF models. Our contributions in this paper are outlined as follows:

\begin{itemize}
    \item We propose a novel architecture, \textit{GLinear}, that can deliver highly accurate TSF by leveraging periodic patterns, making it a promising solution for diverse forecasting applications.
    \item We rigorously perform experiments to validate the \textit{GLinear} architecture through empirical experiments comparing its performance against state-of-the-art methods.
    \item We explore \textit{GLinear}'s applicability across diverse sectors, assessing its impact on forecasting performance with varying data input length and prediction horizons. 
\end{itemize}

The remainder of the paper is organized as follows: Section~\ref{sec:related} presents related work on TSF using state-of-the-art transformer-based models and linear predictors, along with their limitations. Section~\ref{sec:linearmethod} presents the architecture of different linear models to provide a better understanding of the proposed method. Section~\ref{sec:glinear} presents the proposed GLinear model. Section~\ref{sec:setup} presents the details of the experimental setup, including the used datasets, implementation details, and evaluation metrics. Different experiments and their results are presented in Section~\ref{sec:results}. Finally, Section~\ref{sec:conclusion} concludes our research with takeaways and future directions.

\section{Related Work}
\label{sec:related}
\input{sections/relatedwork}

\section{Methodology}
\label{sec:glinear}
\input{sections/proposedmethod}

\section{Experimental Setup}
\label{sec:setup}
\input{sections/experiments}

\section{Experiments and Results}
\label{sec:results}
\input{sections/results}

\section{Conclusion and Future Directions}
\label{sec:conclusion}
\input{sections/conclusion}


\vspace{\baselineskip}

\bibliographystyle{elsarticle-num}
\bibliography{bib2}

\input{bio}

\end{document}

%% file: sections/relatedwork.tex
TSF has become increasingly crucial due to its wide range of real-world applications. Consequently, diverse methodologies have been developed to improve prediction accuracy and robustness. Recent research has explored both simpler (traditional) and complex (DL-based)  approaches. One prominent direction leverages the power of multi-head attention mechanisms within Transformer architectures to capture intricate temporal dependencies~\cite{transformerinfluenza, Informer, Autoformer}. In contrast, other studies~\cite{RDLinear, zeng2023transformers,li2023revisiting} have demonstrated the effectiveness of simpler, computationally efficient models, such as single-layer linear models, for certain forecasting tasks. 
 
\subsection{State-of-the-Art Transformers}
\label{sec:transformer}

Transformer architectures have demonstrated remarkable potential in different DL tasks by effectively capturing long-range dependencies, a critical aspect often overlooked by traditional methods~\cite{wen2022transformers, kanwal2024equipping, kanwal2022attention}. However, adapting Transformers for time series requires addressing inherent challenges such as computational complexity and the lack of inherent inductive biases for sequential data~\cite{transformerinfluenza,Fedformer}. This efficiency bottleneck is addressed with the Informer~\cite{Informer} model with the introduction of ProbSparse attention, which reduces complexity from $O(L^2)$ to $O(L*log L)$ and enables efficient processing of long sequences~\cite{Informer}. They also employed a generative decoder, predicting long sequences in a single forward pass. Another version of the transformer model, Autoformer~\cite{Autoformer}, was proposed to tackle the same complexity issue by replacing an auto-correlation mechanism with the dot product attention to efficiently extract dominant periods in the time series. Their approach proved particularly effective for long-term forecasting on datasets like ETTh and Electricity Transformer~\cite{Autoformer}. Furthermore, Wu \emph{et al.}~\cite{Timesnet} incorporated time-specific inductive biases in their approach. Their proposed model, TimesNet, was introduced to treat time series as images and leverage 2D convolution operations across multiple time scales to capture intra- and inter-variable relationships, achieving state-of-the-art results on various long-term forecasting benchmarks ~\cite{Timesnet}. 

\begin{figure}[!t]
  \includegraphics[width=0.5\textwidth]{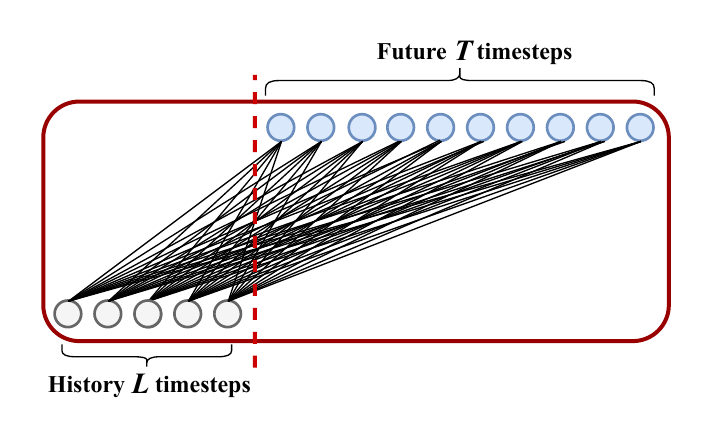}
  \caption{The architecture of LTSF-Linear.}
  \label{lstf}
\end{figure}
The flexibility of Transformer-based models has facilitated their application across diverse forecasting horizons and domains. For short- and medium-term forecasting, adaptations focusing on computational efficiency and local pattern extraction have been explored. For instance, FEDformer~\cite{Fedformer} proposed frequency-enhanced attention and a mixture of expert decoders to capture both global and local patterns efficiently. This approach has shown promising results in short-term load forecasting and other applications where capturing high-frequency components is crucial. For long-term forecasting, the ability of Transformers to model long-range dependencies becomes paramount. Times-Net has demonstrated remarkable performance in this domain~\cite{Timesnet}. Furthermore, some recent research~\cite{Probabilistic, Probabilistic_case} has incorporated external factors and contextual information into Transformer models. Such as integrating weather data or economic indicators to improve forecasting accuracy in domains like energy consumption and financial markets. Additionally, probabilistic forecasting using Transformers is gaining traction, providing not only point predictions but also uncertainty quantification, which is essential for risk management in various applications~\cite{Probabilistic, Probabilistic_case}.

\subsection{State-of-the-Art Linear Predictors}
\label{sec:linearpred}

While Transformer architectures have demonstrated remarkable success, their substantial computational demands and memory footprint pose challenges for deployment in resource-constrained environments, such as edge devices~\cite{kanwal2023vision}. This computational burden has motivated a resurgence of interest in simpler, more efficient models, particularly linear predictors, which offer a compelling balance between forecasting accuracy and computational cost~\cite{zeng2023transformers,li2023revisiting}. Research in this area can be broadly categorized into two main directions: enhancements to traditional linear models through advanced techniques and the development of novel, specialized linear architectures designed explicitly for time series data~\cite{RDLinear, ni2023mixture}.

\begin{figure*}[!t]
  \centering
  \includegraphics[width=\textwidth]{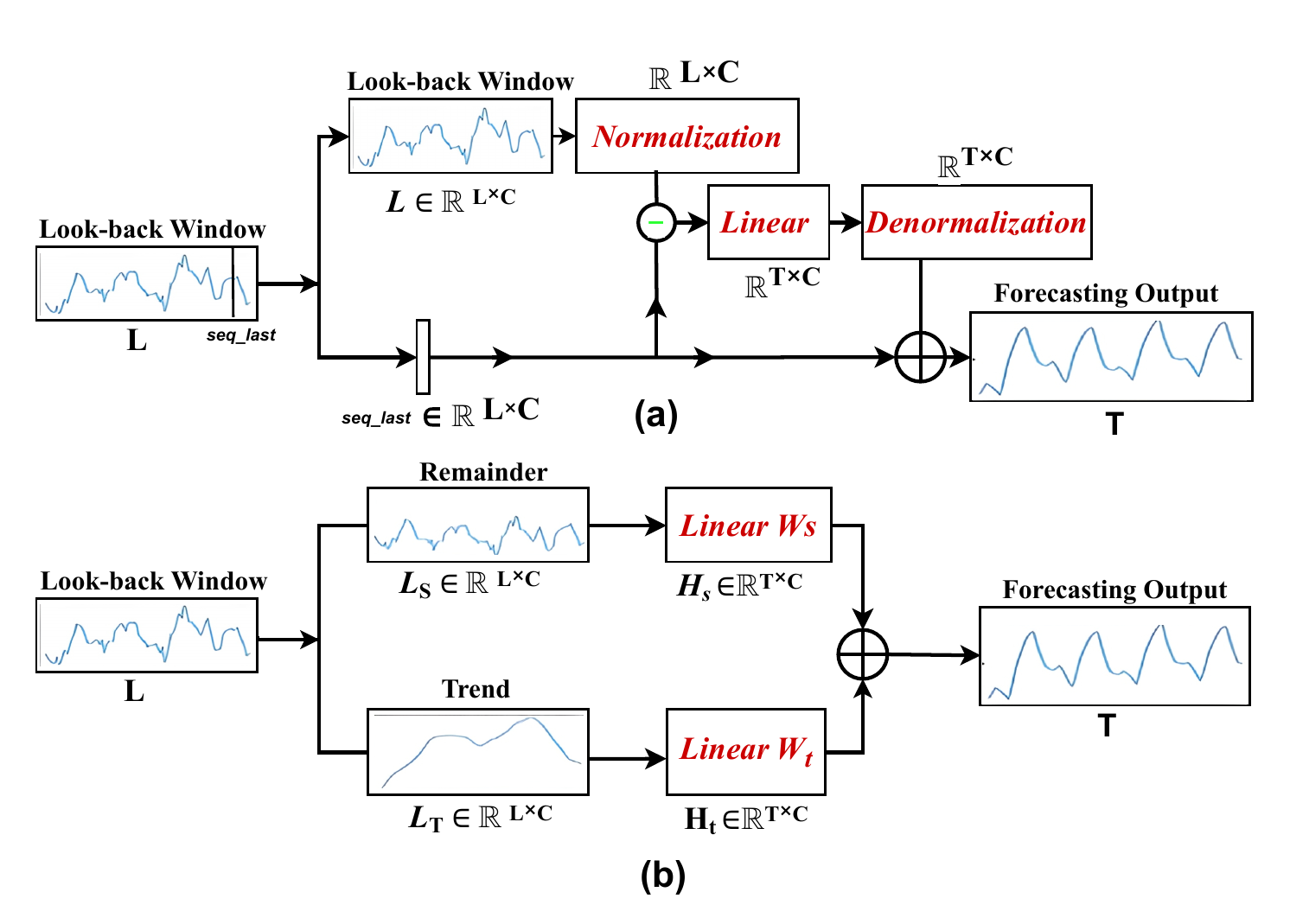}
  \caption{(a) The architecture of the NLinear predictor,  (b) The architecture of the DLinear predictor. The illustration of the architecture is created based on the description in the original paper~\cite{zeng2023transformers}.    }
  \label{fig:NDlinear}
\end{figure*}

One prominent research direction focuses on enhancing classical linear methods to better capture complex temporal dynamics. Traditional methods like auto-regressive (AR)~\cite{Autoregressive} models and their variants, while computationally efficient, often struggle with non-linear patterns and long-range dependencies. Zheng \emph{et al.}~\cite{Dynamic_Regression} introduced a dynamic regression technique that allows the model coefficients to vary over time using adaptive filtering. This approach dynamically adjusts model parameters based on incoming data, improving the adaptability of these models to changing time series characteristics~\cite{Dynamic_Regression}. Other techniques utilizing Kalman filtering within a linear framework have demonstrated effectiveness in tracking evolving trends and seasonality~\cite{extended_Kalman_filter}. Furthermore, it has been shown that the use of sparse linear models, such as LASSO regression, which select only the most relevant past observations for prediction, enhances both efficiency and interpretability~\cite{OBrien2016}. These advancements aim to maximize the performance of established linear frameworks by integrating sophisticated techniques to mitigate their inherent limitations. Another research direction involves developing novel linear architectures specifically tailored for time series data. This includes models like DLinear~\cite{zeng2023transformers}, which decomposes the time series into trend and seasonal components and models them with simple linear layers, achieving surprisingly strong performance on long-term forecasting tasks~\cite{zeng2023transformers}. Similarly, NLinear~\cite{zeng2023transformers} proposes a simple neural network with a single linear layer for forecasting, demonstrating competitive results while drastically reducing computational complexity.

Although linear predictors like NLinear~\cite{zeng2023transformers}, DLinear~\cite{zeng2023transformers}, and RLinear~\cite{li2023revisiting} have achieved competitive accuracy with drastically reduced computational overhead, these models still exhibit some limitations. Such as struggling with capturing complex non-linear patterns or failing to effectively model specific time series characteristics, such as strong seasonality or abrupt changes in trend. Furthermore, these models also rely heavily on extensive historical data to achieve high prediction accuracy. This dependency can limit their effectiveness in scenarios with limited data availability, such as newly established systems or rapidly changing environments. These issues motivate the development of novel architectures and improvements in linear models, which aim to address these shortcomings. 

\textbf{\textit{GLinear}} achieves superior results by integrating two components: (1) a non-linear \emph{ Gaussian Error Linear Unit (GELU)-based transformation layer} to capture intricate patterns and (2) \emph{Reversible Instance Normalization (RevIN)} to standardize data distributions across instances, ensuring consistent performance and adaptability across diverse datasets. This approach provides a more comprehensive and efficient solution for TSF.

\section{Architecture of Different Linear Models}
\label{sec:linearmethod}
\input{sections/linearmethod}

%% file: sections/linearmethod.tex
Different state-of-the-art linear predictors are explained in this section to contrast enhancements of the proposed GLinear~\cite{zeng2023transformers}. 

The input of a time series predictor is the $L$ time steps of past input samples (also referred to as the \emph{lookup window} or \emph{input sequence length}). A predictor uses this input to predict $T$ time steps of future values (also referred to as \emph{output horizon} or \emph{prediction length}), and $C$ refers to channels. The details of different linear predictors are given below:

\subsection{Linear Predictor (LTSF-Linear)}
The first model is a \emph{LTSF-Linear} predictor~\cite{zeng2023transformers} where LTSF stands for long-term TSF. The architecture of the linear predictor can be visualized from Figure~\ref{lstf}. This predictor is composed of a single fully connected linear layer, or dense layer. LTSF-Linear predictor does not model any spatial correlations. A single temporal linear layer directly regresses historical time series for future prediction via a weighted sum operation, as follows:~\footnote{Notation: Scalars are denoted by italic letters, vectors by bold lowercase (e.g., $\mathbf{x}$), and matrices by bold uppercase (e.g., $\mathbf{W}$). The input sequence length is $L$, the prediction length is $T$, the number of channels is $C$, and the batch size is $B$.}

\begin{equation} \label{eq:1}
\mathbf{T} = \mathbf{W}\,\mathbf{L} + \mathbf{b},
\quad
\mathbf{W}\in\mathbb{R}^{T\times L},\;
\mathbf{L}\in\mathbb{R}^{L},\;
\mathbf{b},\mathbf{T}\in\mathbb{R}^{T}.
\end{equation}
For simplicity, Eq.~\eqref{eq:1} is written in the univariate form. In the multivariate case (with $C$ channels) the same linear mapping is applied independently per channel, i.e. for each $c\in\{1,\dots, C\}$ we have $\hat{\mathbf{x}}^{(c)}=\mathbf{W}\,\mathbf{x}^{(c)}+\mathbf{b}$, yielding $\hat{\mathbf{X}}\in\mathbb{R}^{T\times C}$. For batched inputs, $\mathbf{X}\in\mathbb{R}^{B\times L\times C}$ the mapping is applied over the $L$ $\to$ $T$ axis for every $(b,c)$ pair.

\subsection{NLinear Predictor (Normalization-based Linear Model)}

To boost the performance of LTSF-Linear, \emph{NLinear}~\cite{zeng2023transformers} performs normalization to tackle the distribution shift in the dataset. NLinear first subtracts the input by the last value of the sequence, as shown in Figure~\ref{fig:NDlinear} (a). Then, the input goes through a linear layer, and the subtracted part is added back before making the final prediction. The subtraction and addition in NLinear are a simple normalization for the input sequence.

\subsection{DLinear Predictor (Decomposition-based Linear Model)}
\emph{DLinear}~\cite{zeng2023transformers} is a combination of a decomposition scheme used in Autoformer and FEDformer with linear layers. It first decomposes raw data input into a trend component by a moving average kernel and a remainder (seasonal) component, as shown in Figure~\ref{fig:NDlinear} (b). Then, two one-layer linear layers are applied to each component, and the two features are summed up to get the final prediction. By explicitly handling trend, DLinear enhances the performance of a vanilla linear model when there is a clear trend in the data.

\subsection{RLinear Predictor (Reversible normalization-based Linear Model)}
\begin{figure}[h!]
  \centering
  \includegraphics[width=0.2\textwidth]{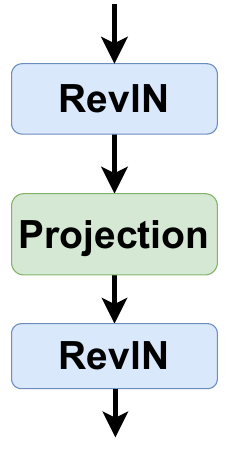}
  \caption{The schematic of the architecture of RLinear. The illustration of the architecture is created based on the description in the original paper~\cite{li2023revisiting}.}
  \label{fig:rlinear}
\end{figure}

\emph{RLinear}~\cite{li2023revisiting} combines a linear projection layer with \textit{RevIN} to achieve competitive performance, as shown in Figure~\ref{fig:rlinear}. The study reveals that RevIN enhances the model's ability to handle distribution shifts and normalize input data effectively, leading to improved results even with a simpler architecture.

%% file: sections/proposedmethod.tex
Existing linear models use different simple operations like normalization and decomposition. It is worth noting that with the involvement of simpler mathematical operations, it is possible to build powerful linear predictors with some variations of activation functions to extract meaningful results for the required task~\cite{ni2023mixture}. Keeping these in mind, a new Gaussian-activated linear \textbf{\emph{GLinear}} predictor is proposed.

\begin{figure}[!ht]
  \centering
  \includegraphics[width=0.65\columnwidth]{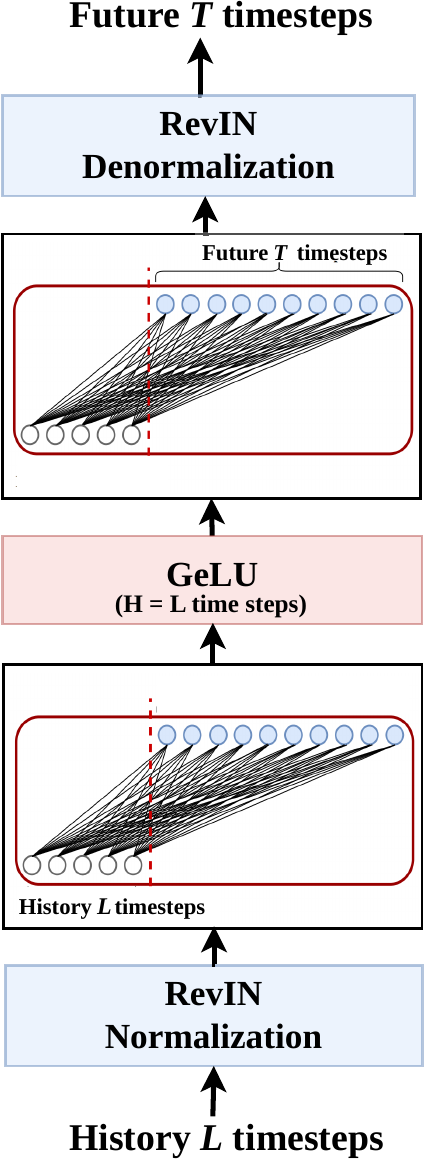}
  \caption{The architecture of the proposed GLinear predictor.}
  \label{Glinear}
\end{figure}

Figure~\ref{Glinear} shows the architecture of the GLinear predictor that is composed of two fully connected layers of the same input size, having a GELU nonlinearity in between them. Different configurations of input and layer sizes are tested to lead to this final architecture. 

Furthermore, \emph{RevIN} is applied to the input and output layers of the GLinear model~\cite{kim2021reversible}. This normalization layer transforms the original data distribution into a mean-centered distribution, where the distribution discrepancy between different instances is reduced. This normalized data is then applied as new input to the used predictor, and then the final output is de-normalized at the last step to provide the final prediction. RevIN can be used with any existing model. It does not add any significant overhead in training time due to the low computational complexity of this normalization scheme. 

The time-series data usually suffers from a distribution shift problem due to changes in its mean and variance over time that can degrade the performance of time-series predictors. RevIN utilizes a learnable affine transformation to remove and restore the statistical information of a time-series instance that can be helpful to handle datasets with distribution shift problems. 

\subsection{Mathematical Formulation of GLinear}
\label{sec:glinear_math} 

Let the input batch be $\mathbf{X}\in\mathbb{R}^{B\times L\times C}$,
where $B$ is the batch size, $L$ is the lookup window (input sequence length),
and $C$ is the number of variables (channels).
The forecasting objective is to predict $\widehat{\mathbf{Y}}\in\mathbb{R}^{B\times T\times C}$,
where $T$ is the output horizon (prediction length).

\subsubsection{Reversible Instance Normalization (RevIN)}
For each instance $b$ and channel $c$, we compute the per-instance statistics
\begin{equation}
    \mu_{b,c} \;=\; \frac{1}{L}\sum_{t=1}^{L} X_{b,t,c}, \qquad
    \sigma_{b,c} \;=\; \sqrt{\frac{1}{L}\sum_{t=1}^{L}\big(X_{b,t,c}-\mu_{b,c}\big)^2 + \varepsilon }.
\end{equation}
The normalized (and affine-transformed) input becomes
\begin{equation}
   \tilde{X}_{b,t,c} \;=\; \gamma_c\odot\frac{X_{b,t,c}-\mu_{b,c}}{\sigma_{b,c}} + \beta_c,
    \qquad
\widetilde{\mathbf{X}}\in\mathbb{R}^{B\times L\times C}.
\end{equation}
where \(\gamma_c,\beta_c\) are learnable per-channel affine parameters. For processing, we reshape each instance/channel to a column vector:
\[
\mathbf{x}^{(b,c)} = \big(\widetilde{X}_{b,1,c}, \dots, \widetilde{X}_{b,L,c}\big)^{\!\top}\in\mathbb{R}^{L}.
\]

\subsubsection{Gaussian Distribution-based Non-linear Mapping}
After reshaping, each channel $c$ is processed independently (channel-wise mapping). For every batch index $b$ and channel $c$, we apply a compact two-layer MLP with the GELU nonlinearity. Let the (per-instance, per-channel) input column vector be
\(\mathbf{x}^{(b,c)}\in\mathbb{R}^{L}\), we also fixed $\vect{H}=\vect{L}$ in the experiments.


\begin{align}
\mathbf{h}^{(b,c)} &= \mathrm{GELU}\!\big(\mathbf{W}^{(1,c)}\mathbf{x}^{(b,c)} + \mathbf{b}^{(1,c)}\big), \label{eq:glinear_h}\\
&\quad \mathbf{W}^{(1,c)}\in\mathbb{R}^{L\times L},\; \mathbf{b}^{(1,c)}\in\mathbb{R}^{L}, \nonumber\\
\mathbf{o}^{(b,c)} &= \mathbf{W}^{(2,c)}\mathbf{h}^{(b,c)} + \mathbf{b}^{(2,c)}, \label{eq:glinear_o}\\
&\quad \mathbf{W}^{(2,c)}\in\mathbb{R}^{T\times L},\; \mathbf{b}^{(2,c)}\in\mathbb{R}^{T}. \nonumber
\end{align}
Here $\mathrm{GELU}$~ is the Gaussian Error Linear Unit~\cite{hendrycks2016gaussian},  a smooth non-linear activation function, defined as,
\begin{equation}
    \mathrm{GELU}(x) = x \cdot \Phi(x),
    \qquad \texttt{where} \quad
    \Phi(x) = \tfrac{1}{2}\Big(1 + \mathrm{erf}\!\big(\tfrac{x}{\sqrt{2}}\big)\Big).
\end{equation}
For fast approximation, we adopted GELU using \[ \mathrm{GELU}(x) \approx 0.5 x \Big(1 + \tanh\big(\sqrt{\tfrac{2}{\pi}}(x + 0.044715 x^3)\big)\Big). \]
\subsubsection{Denormalization}
After the model outputs \(\mathbf{o}^{(b,c)}\in\mathbb{R}^T\), we reverse the affine and instance scaling:
\begin{equation}
    \widehat{Y}_{b,t,c} \;=\; \frac{o^{(b,c)}_t - \beta_c}{\gamma_c}\cdot\sigma_{b,c} + \mu_{b,c}, 
    \qquad
    \widehat{\mathbf{Y}}\in\mathbb{R}^{B\times T\times C}.
\end{equation}


\subsection{Features of GLinear}
Some features of the GLinear model are
\begin{itemize}
\item It is a simpler model; it is not made up of any complex components, functions, or blocks (like self-attention schemes, positional encoding blocks, etc.). It integrates two components: (1) a non-linear GELU-based transformation layer to capture intricate patterns and (2) Reversible Instance Normalization (RevIN).

\item Due to its simple architecture, training of this model is significantly faster than transformer-based predictors. The parameter count per channel is
\begin{equation}
    |W^{(1)}| + |b^{(1)}| + |W^{(2)}| + |b^{(2)}|
    = L^2 + L + TL + T.
\end{equation}

With $C$ channels (independent weights), this scales as
$O(C \cdot (L^2 + TL))$.
This is significantly more efficient than transformer-based models,
which scale as $O(L^2 C)$ in the self-attention layers.

GLinear has the potential to provide competitive performance to other state-of-the-art linear predictors over different benchmarks.
Unlike NLinear, which only tackles distribution shift, and DLinear, which requires explicit trend decomposition, GLinear introduces smooth Gaussian nonlinearity within a lightweight linear framework. This enables it to capture nonlinear seasonality while retaining the computational efficiency of linear predictors, bridging the gap between purely linear mappings and heavy transformer-based models.

\end{itemize}

%% file: sections/experiments.tex
In this section, we present the details about the datasets used, the experimental setup, and the evaluation metrics.

\subsection{Dataset}
We conducted experiments on four different real-world datasets (ETTh1, Electricity, Weather, and Traffic). Table~\ref{tab1} provides a brief overview of these datasets.

\subsubsection{ETTh1}
The ETTh1 dataset is used for long-sequence TSF in electric power. It includes two years of data from two Chinese counties, focusing on Electricity Transformer Temperature (ETT)~\cite{haoyietal-informer-2021}. It's designed for detailed exploration of forecasting problems. This dataset is crucial for analyzing transformer temperatures and power load features in the electric power sector. ETTh1 differs from ETTh2 in granularity, focusing on long sequences compared to ETTh2's hourly forecasting. The ETTh dataset serves the purpose of aiding research and analysis in the electric power sector, particularly for forecasting transformer temperatures and power load features. Applications of the ETTh dataset include research in long sequence time-series forecasting and studying power load features for better power deployment strategies.

The ETTh1 dataset is a multivariate time series dataset having 7 different variables (channels). It contains data of 725.83 days with a granularity of 1 hour, meaning each timestamp represents a one-hour interval of data that provides 17420 timestamp values of each variable (17420 / 24 hours per day = 725.83 days of data). 

\subsubsection{Electricity}

The Electricity dataset~\cite{khan2020towards} is also a multivariate time series dataset having 321 channels. It contains data of 1096 days with a granularity of 1 hour, which provides 26304 timestamp values of each channel (26304 / 24 hours per day = 1096 days of data).

\subsubsection{Weather}

The Weather dataset~\cite{angryk2020multivariate} contains 52696 timestamp values collected in 365.86 days; each timestamp has 21 channels and a granularity of 10 minutes. 

\subsubsection{Traffic}
The Traffic dataset~\cite{chen2001freeway} contains 731 days of data with a granularity of 1 hour. It provides data from 862 channels, each having 17,544 timestamp values. 

\begin{table}[h]
\caption{Overview of ETTh1, Electricity, Weather, and Traffic datasets.}
\begin{center}
\resizebox{0.99\columnwidth}{!}{%
\begin{tabular}{|c|c|c|c|}
\hline
\textbf{}&\textbf{}&\textbf{}&\textbf{} \\
\textbf{Datasets}&\textbf{Timestamps} &\textbf{Variables (Channels)} &\textbf{Granularity} \\
\textbf{}&\textbf{}&\textbf{}&\textbf{} \\

\hline

ETTh1& 17420  & 7 &  1 hour  \\
Electricity& 26304 & 321 &  1 hour  \\
Weather & 52696 & 21&  10 minutes \\
Traffic &17544   & 862 & 1 hour   \\
\hline
\end{tabular}}
\label{tab1}
\end{center}
\end{table}
\subsection{Implementation Details}

The GLinear model is implemented using Python and is sourced from the official PyTorch implementation of LTSF-Linear~\footnote{\url{https://github.com/cure-lab/LTSF-Linear/}}. The respective code repositories contain the training and evaluation protocols of Autoformer, NLinear, and DLinear. Similarly, the RLinear~\footnote{\url{https://github.com/plumprc/RTSF/blob/main/models/RLinear.py}} model is trained and evaluated using the same protocol to ensure a fair comparison across all models.

The same set of hyperparameters is used for training all linear models, such as using the Mean Squared Error (MSE) criterion,
the Adam~\cite{adam_optimizer} optimizer, and a learning rate of 0.001. The source code is publicly available on \href{https://github.com/t-rizvi/GLinear/}{GitHub}.

\subsection{Evaluation Metrics}
The evaluation metrics used for comparison are MSE and Mean Absolute Error (MAE). These metrics are commonly used to assess the accuracy and performance of predictors~\cite{zeng2023transformers, li2024deep}.

%% file: sections/results.tex
\begin{table*}[ht!]
\label{tab:table1}
\centering
\caption{Multivariate forecasting performance comparison across the datasets and models (lower MSE/MAE is better). The input sequence length for all experiments is 336 time steps, and a range of prediction lengths \{12, 24, 48, 96, 192, 336, 720\} is used. The top-performing results are marked in \textbf{bold}. The second-best results are \underline{underlined}.}
\resizebox{1.01\textwidth}{!}{
\begin{tabular}{ |p{1.7cm}||p{0.15cm}|p{1.05cm}|p{1.05cm}|p{1.05cm}||p{1.05cm}|p{1.05cm}|p{1.05cm}|p{1.1cm}||p{1.05cm}|p{1.15cm}|p{1.05cm}| }
 \hline
\multicolumn{2}{|c|}{} & \multicolumn{10}{|c|}{\textbf{\shortstack{Lookup Window - Learning Rate 0.001\\ (Input Sequence Length) = 336}}} \\
 \hline
\multicolumn{2}{|c|}{\textbf{Methods}} & \multicolumn{2}{|c|}{\textbf{Autoformer}} & \multicolumn{2}{|c|}{\textbf{NLinear}} & \multicolumn{2}{|c|}{\textbf{DLinear}} &\multicolumn{2}{|c|}{\textbf{RLinear}} &\multicolumn{2}{|c|}{\textbf{GLinear }}   \\
 \hline
 \multicolumn{2}{|c|}{\shortstack{Dataset/Output Hori.\\ (Prediction Length)}} & MSE & MAE & MSE & MAE & MSE & MAE &  MSE & MAE & MSE & MAE  \\
 \hline
  \multirow{7}{*}{Electricity} & 12 &  0.1638 &	0.2872 & 0.1000	&0.2006	&0.0997	&0.2009	& \underline{0.0967}	&0.1973	& \textbf{0.0883}	&0.1860  \\

 & 24 & 0.1711&	0.2917&	0.1103&	0.2092&	0.1099&	0.2089&	\underline{0.1060}&	0.2049&	\textbf{0.0988}&	0.1952  \\
& 48&	0.1827&	0.2990&	0.1255&	0.2232&	0.1249&	0.2231& \underline{0.1201}&	0.2180&	\textbf{0.1144}&	0.2101\\
&96&	0.1960&	0.3106&	0.1409&	0.2366&	0.1401&	0.2374&	\underline{0.1358}&	0.2317&	\textbf{0.1313}&	0.2258\\
&192&	0.2064&	0.3182&	0.1551&	0.2488&	0.1538&	0.2505&	\underline{0.1518}	&0.2455	& \textbf{0.1494}&	0.2423\\
&336&	0.2177&	0.3290&	0.1717&	0.2654&	0.1693&	0.2678&	\underline{0.1688}&	0.2621&	\textbf{0.1651}&	0.2582\\
&720&	0.2477&	0.3528&	0.2104&	0.2977&	\underline{0.2042}&	0.3005&	0.2071&	0.2940&	\textbf{0.2027}&	0.2906\\

 \hline
 \multirow{7}{*}{ETTh1} & 12&	0.3991&	0.4422&	0.3069&	0.3564&	0.2976&	0.3494&	\underline{0.2862}&	0.3412&	\textbf{0.2848}&	0.3448\\
&24&	0.4759&	0.4733&	0.3474&	0.3842&	0.3194&	0.3627&	\textbf{0.3090}&	0.3559&	\underline{0.3142}&	0.3654\\
&48	&0.5046&	0.4831&	0.3553&	0.3845&	0.3477&	0.3803&	\textbf{0.3454}&	0.3766&	\underline{0.3537}&	0.3869\\
&96	&0.5392&	0.4979&	\underline{0.3731}&	0.3941&	\textbf{0.3705}&	0.3919&	0.3901&	0.4054&	0.3820&	0.4025\\
&192&	0.4907&	0.4906&	\underline{0.4089}	&0.4157&	\textbf{0.4044}&	0.4128&	0.4223&	0.4279&	0.4202&	0.4269\\
&336&	0.4805&	0.4886&	\textbf{0.4324}	&0.4307&	0.4553&	0.4582&	\underline{0.4417}&	0.4383&	0.4915&	0.4715\\
&720&	0.6303&	0.5930	& \textbf{0.4369}&	0.4527&	0.4975&	0.5087&	\underline{0.4634}&	0.4686&	0.5923	&0.5372\\
 \hline
 \multirow{7}{*}{Traffic} & 12&	0.5624&	0.3830&	0.3623&	0.2662&	\underline{0.3610}&	0.2644&	0.3762	&0.2744&	\textbf{0.3222}&	0.2385\\
&24&	0.5801	&0.3786	&0.3719	&0.2682&	\underline{0.3709}&	0.2672&	0.3834&	0.2775&	\textbf{0.3369}&	0.2471\\
& 48	&0.6060	&0.3796	&0.3945&	0.2769&	\underline{0.3932}	&0.2760&	0.4041	&0.2864&	\textbf{0.3630} &	0.2607\\
&96&	0.6426&	0.3998&	0.4113&	0.2820&	\underline{0.4104}&	0.2829&	0.4194&	0.2921&	\textbf{0.3875}&	0.2718\\
&192&	0.6425&	0.3967&	0.4245&	0.2872&	\underline{0.4229}&	0.2881&	0.4323&	0.2965&	\textbf{0.4056}&	0.2802\\
&336&	0.6675&	0.4088	&0.4375&	0.2943&	\underline{0.4362}&	0.2961&	0.4451&	0.3027&	\textbf{0.4200}&	0.2871\\
&720&	0.6570	&0.4030&	\underline{0.4657}&	0.3109&	0.4660&	0.3152&	0.4733&	0.3191&	\textbf{0.4488}&	0.3038
  \\
 \hline
 \multirow{7}{*}{Weather} & 12 & 	0.2010	 & 0.2933 & 	0.0784 & 	0.1127 & 	0.0783 & 	0.1158 & 	\textbf{0.0706} & 	0.0974 & 	\underline{0.0716} & 	0.0940\\
 & 24 & 	0.2095	 & 0.3033 & 	0.1056	 & 0.1453 & 	0.1040	 & 0.1519 & 	\textbf{0.0905} & 	0.1247 & 	\underline{0.0909} & 	0.1247\\
 & 48 & 	0.2397 & 	0.3202 & 	0.1357 & 	0.1824 & 	0.1367	 & 0.1937 & \textbf{0.1138}	 & 0.1566	 & \underline{0.1163} & 	0.1602\\
 & 96 & 	0.3004 & 	0.3776	 & 0.1761 & 	0.2264	 & 0.1756 & 0.2386 & 	\textbf{0.1450}	 & 0.1936	 & \underline{0.1457}	 & 0.1966\\
 & 192 & 	0.3916	 & 0.4382 & 	0.2164 & 	0.2595	 & 0.2160 & 	0.2739 & 	\textbf{0.1878} & 	0.2339 & 	\underline{0.1883} & 	0.2385\\
 & 336 & 	0.3830 & 	0.4171 & 	0.2664 & 	0.2966 & 	0.2652	 & 0.3192	 & \textbf{0.2404}	 & 0.2743	 & \underline{0.2407}	 & 0.2764\\
 & 720 & 	0.5420	 & 0.5032	 & 0.3339	 & 0.3437 & 	0.3275 & 	0.3667 & 	\textbf{0.3159}	 & 0.3271 & 	\underline{0.3200}	 & 0.3334\\
 \hline
 \multicolumn{2}{|c|}{\textbf{Top 1 Performing }} & \multicolumn{2}{|c|}{0} & \multicolumn{2}{|c|}{2} & \multicolumn{2}{|c|}{2} &\multicolumn{2}{|c|}{9} &\multicolumn{2}{|c|}{\textbf{15}}   \\
 \hline
 \multicolumn{2}{|c|}{\textbf{Top 2 Performing }} & \multicolumn{2}{|c|}{0} & \multicolumn{2}{|c|}{5} & \multicolumn{2}{|c|}{10} &\multicolumn{2}{|c|}{17} &\multicolumn{2}{|c|}{\textbf{23}}   \\
 \hline

\end{tabular}}
\label{tab2}
\end{table*}

\begin{figure*}[!t]
\centering
\includegraphics[width=0.495\textwidth]{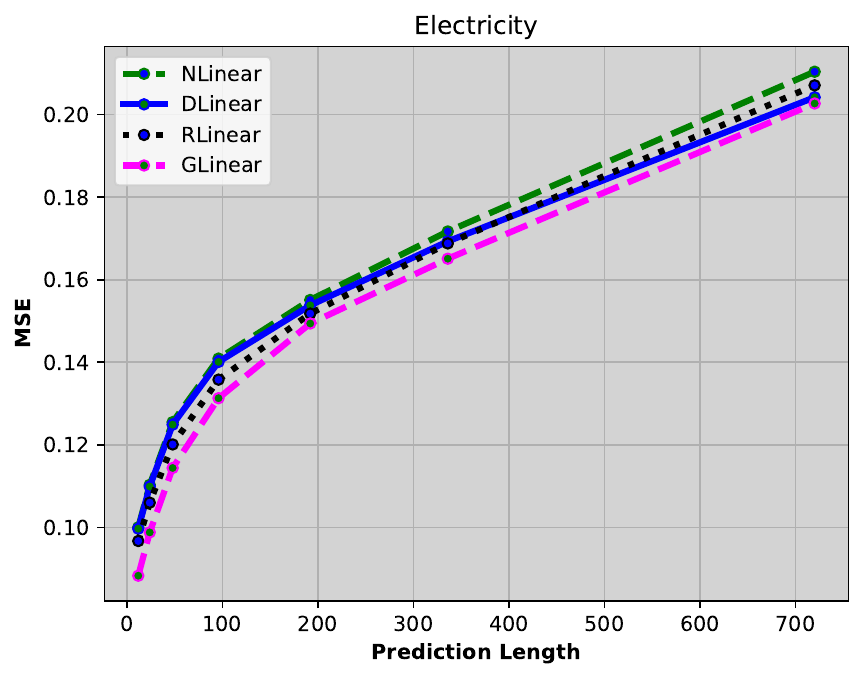}
\hfill
\includegraphics[width=0.495\textwidth]{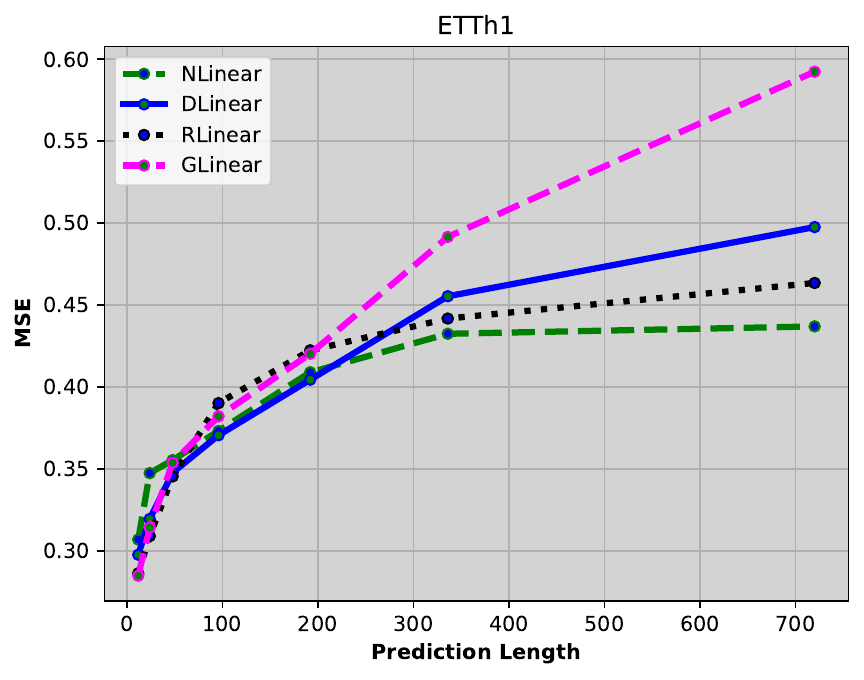}
\includegraphics[width=0.495\textwidth]{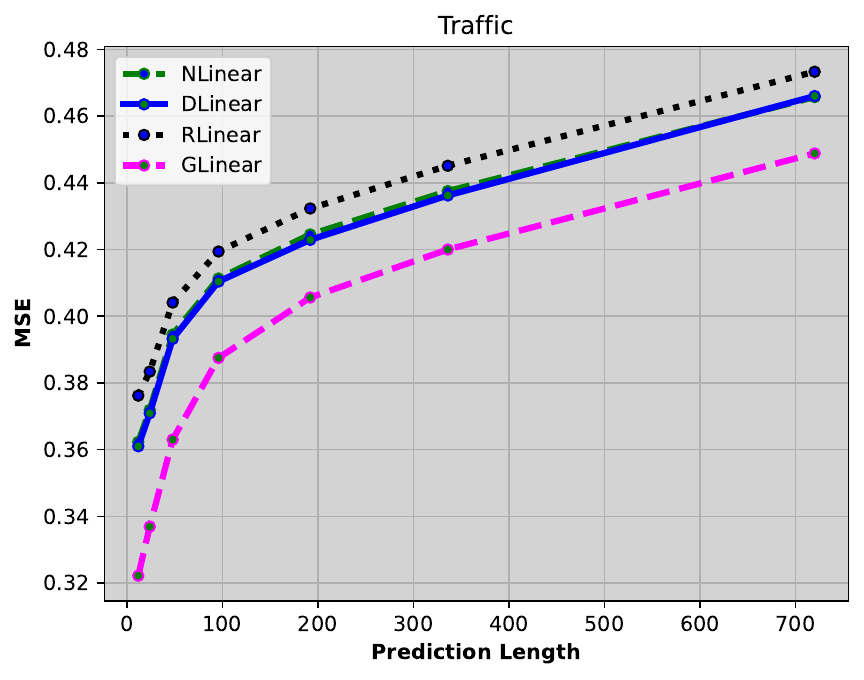}
\hfill
\includegraphics[width=0.495\textwidth]{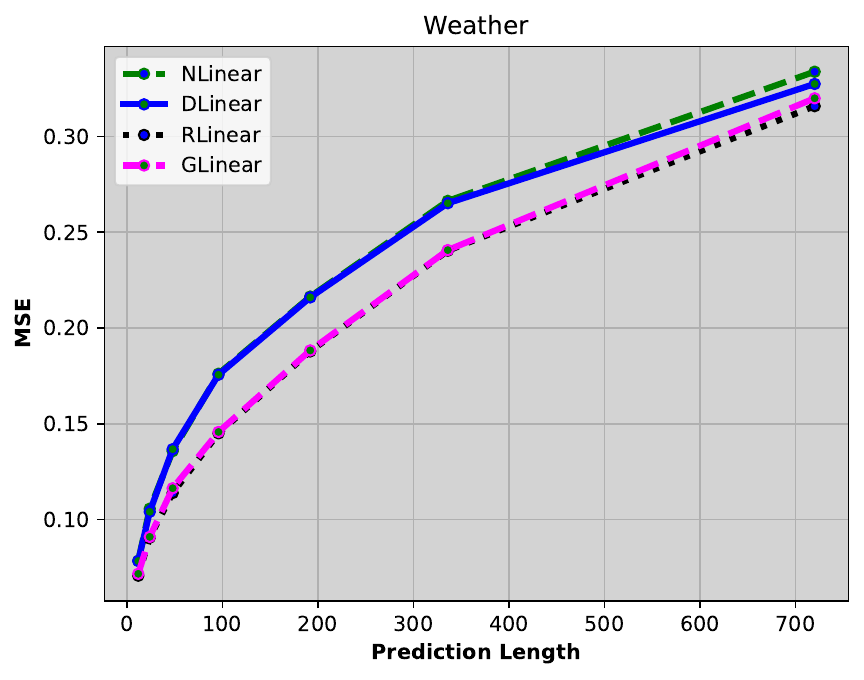}
\caption{Multivariate Forecasting performance comparison across the datasets and the models. Input sequence length is 336, and values of prediction lengths are \{12, 24, 48, 96, 192, 336, 720\}. The datasets are (a) Electricity, (b) ETTh1, (c) Traffic, and (d) Weather.}
\label{res:comparegraph}
\end{figure*}

\begin{table*}[h]
\centering
\caption{\vspace{0.5em}The average per-epoch training time and estimated per-sample inference latency on the Electricity dataset
(Input length = 336, prediction length = 720, batch size = 16). 
Inference latency is estimated from average iteration speed in the training logs.\vspace{0.3em}}
\label{tab:runtime_comparison}
\resizebox{0.65\textwidth}{!}{\begin{tabular}{lcc}
\hline
\textbf{Model} & \textbf{Per-epoch Training Time (s)} & \textbf{ Inference (ms/sample)} \\
\hline
NLinear~\cite{zeng2023transformers} & 21.32   & 2.01 \\
DLinear~\cite{zeng2023transformers}  & 24.75   & 2.27 \\
RLinear~\cite{li2023revisiting} & 233.10  & 15.83 \\
\hline
GLinear (ours) & 23.22 & 2.15 \\
\hline
\end{tabular}}
\end{table*}

The experimental setup was designed to assess GLinear's performance in both short-term and long-term forecasting, as well as to analyze the impact of varying historical data lengths using two different experiments.

For a fair comparison, we have focused on comparing GLinear against other linear models from the literature because these previous works (NLinear~\cite{zeng2023transformers}, DLinear~\cite{zeng2023transformers}, RLinear~\cite{li2023revisiting}) have already been benchmarked against five recent transformer-based methods (Autoformer~\cite{Autoformer}, FEDformer~\cite{Fedformer},  Informer~\cite{haoyietal-informer-2021}, Pyraformer~\cite{pyraformer}, and LogTrans~\cite{logtrans}) and demonstrated that linear models outperform these transformers. To provide context while keeping the experiments concise, we compared GLinear with the linear baselines and Autoformer as a representative transformer.

\subsection{Evaluating Different Prediction Lengths for Fixed Input Length}

In the first experiment, the length of the input sequence was fixed at 336 time steps, representing a historical window to learn the underlying patterns. The prediction lengths were varied across multiple time frames to assess the model’s ability to forecast different horizons, as displayed in Table~\ref{tab2}. Table~\ref{tab2} provides a comprehensive evaluation of all predictors, including NLinear~\cite{zeng2023transformers}, DLinear~\cite{zeng2023transformers}, RLinear~\cite{li2023revisiting}, Autoformer~\cite{Autoformer}, and the proposed GLinear across four datasets. In an extensive series of experiments, we varied the prediction lengths in the range: \{12, 24, 48, 96, 192, 336, 720\}. This set of experiments allows for an evaluation of the model's performance in forecasting short-, medium-, and long-term future steps, helping to gauge the effectiveness of the model for various forecasting time horizons. We used a quantitative approach to score the best-performing candidate in all datasets, giving a point for the best and second-best results in each row. The results conclude that \textbf{\textit{GLinear}} is a majority winner with substantially improved results in both evaluation metrics.

Compared strictly against the linear models (NLinear~\cite{zeng2023transformers}, DLinear~\cite{zeng2023transformers}, and  RLinear~\cite{li2023revisiting}), smooth nonlinearities introduced by the GLinear architecture capture more complex patterns such as seasonality and nonlinear dependencies. By restricting operations to per-channel projections, GLinear preserves temporal locality and avoids spurious inter-channel correlations that often hamper the generalization of other linear models. The insertion of a light nonlinearity (GELU) endows the model with the capacity to approximate smooth nonlinear mappings, which pure linear predictors lack. Together with RevIN, which stabilizes scale shifts across instances, this design biases GLinear toward robust short-to-medium horizon forecasting. These inductive biases explain why GLinear can outperform both heavier Transformer-based models and simpler linear models across diverse datasets. In addition, GLinear’s per‑channel linear mapping with a single smooth nonlinearity, together with RevIN, surfaces component effects and instance‑level scale shifts more directly than deeper attention stacks (as deployed in the timeseries Transformers), which supports interpretability in multivariate settings. By avoiding quadratic attention and large attention blocks typical of time‑series Transformers, GLinear scales favorably in both training and inference; this aligns with Table~\ref{tab:runtime_comparison} and prior works~\cite{zeng2023transformers, li2023revisiting} advocating lightweight mappings for long sequences.

Figure~\ref{res:comparegraph} further compares the forecasting performance of these models. The results demonstrate that the proposed GLinear model outperforms other predictors in most cases (a lower MSE indicates better performance). Specifically, GLinear achieves the highest performance for the Electricity and Traffic datasets. For weather forecasting, GLinear is the second-best model, with RLinear taking the top spot. However, the difference in MSE between the two models is minimal, as shown in Figure~\ref{res:comparegraph} (d). For the ETTh1 dataset, no single model consistently outperforms others across all prediction lengths. For the ETTh1 dataset, GLinear ranks among the top-performing models for shorter prediction lengths (12, 24, and 48), but the best-performing model changes as the prediction length increases. We noted that on ETTh1, GLinear's performance degrades as the horizon lengthens. This effect is expected, since the channel-wise projection does not explicitly model long-range temporal dependencies, making the approximation error accumulate over larger horizons.\\

Table~\ref{tab:runtime_comparison} shows the computational efficiency of the linear models. Overall, all linear predictors (NLinear, DLinear, and the proposed GLinear) except RLinear have comparable training and inference time. However, by comparing the cross-sections, GLinear achieves a better per-epoch training time than DLinear and RLinear, but it has a higher per-epoch training time than NLinear, which has a very simplistic architecture. GLinear achieves relatively lower inference latency comparable to the other linear baselines (RLinear and DLinear), and is approximately 10$\times$ faster in training than RLinear, while maintaining superior forecasting accuracy. The table also shows that RLinear is the slowest, due to its complicated channel-wise linear transformations, but overall yields slightly superior results to NLinear and DLinear.
The above results demonstrate that the GLinear is a promising linear model for real-time scenarios with competitive forecasting performance.

\subsection{Impact of Input Sequence Length on Forecasting Future Steps}

\begin{figure*}[!t]
\centering
\includegraphics[width=0.49\textwidth]{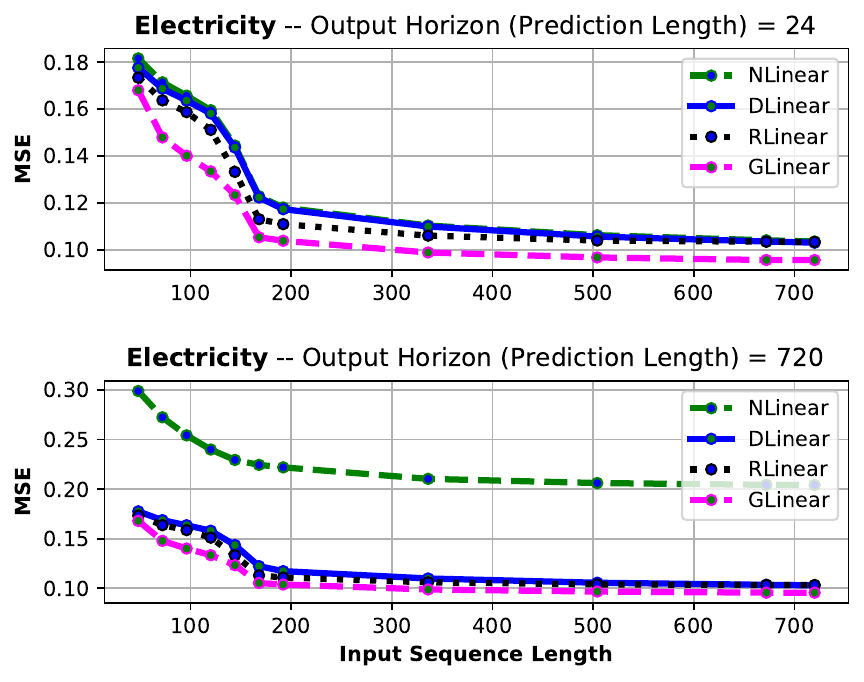}
\hfill
\includegraphics[width=0.49\textwidth]{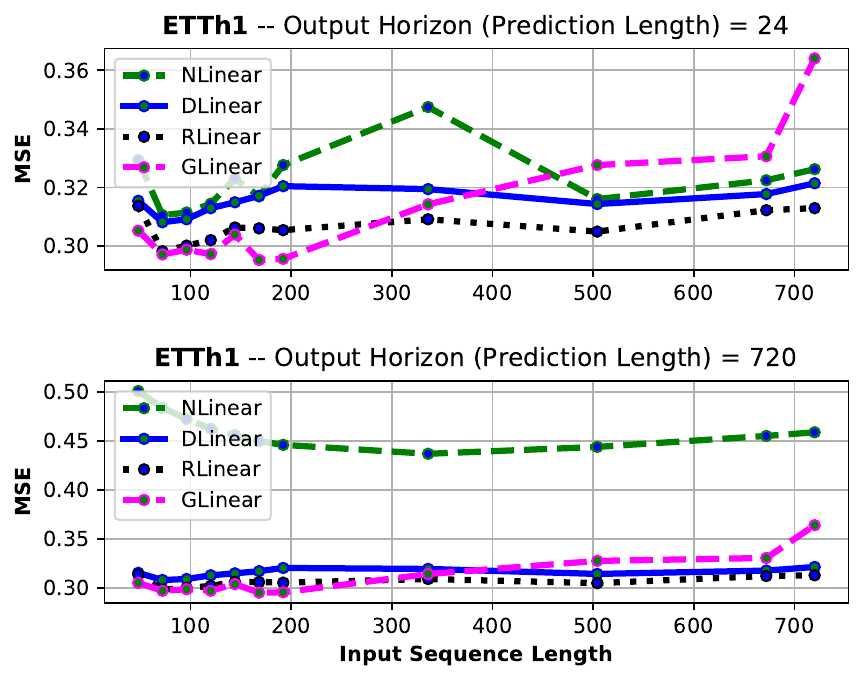}

\medskip

\includegraphics[width=0.49\textwidth]{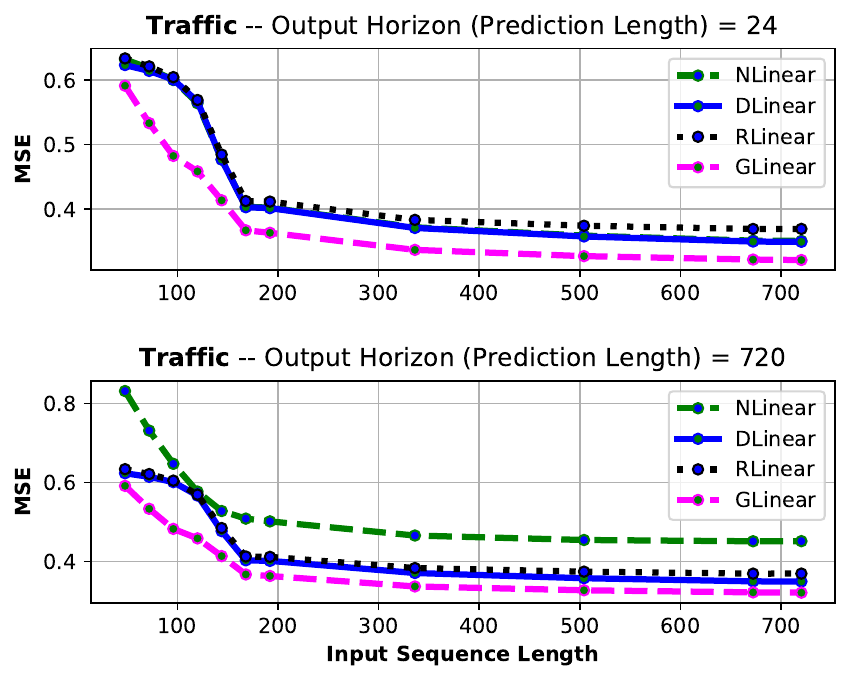}
\hfill
\includegraphics[width=0.49\textwidth]{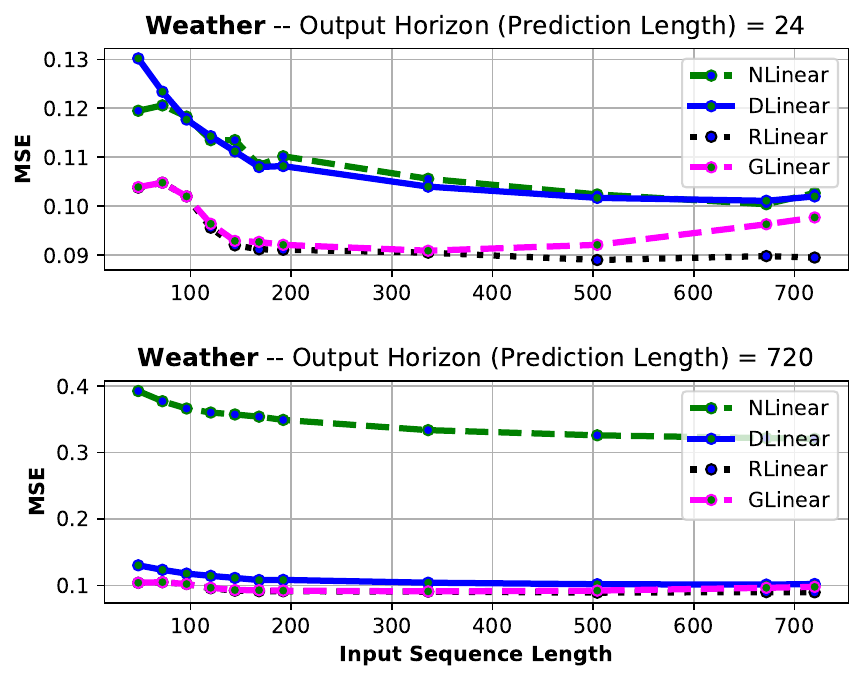}

\caption{A performance comparison of the linear predictors (NLinear, DLinear, RLinear, and the proposed GLinear) with varying input sequence lengths in both long-term forecasting (720 time steps) and short-term forecasting (24 time steps) for different datasets: (a) Electricity, (b) ETTh1, (c) Traffic, and (d) Weather.}
\label{res:combineexp2}
\end{figure*}

In the second experiment, the length of the input sequence was varied to understand how much historical data is required to accurately forecast the 24 and 720 future time steps. The input sequence lengths were set to \{48, 72, 96, 120, 144, 168, 192, 336, 504, 672, 720\}. For each of these input lengths, the model was trained to predict 24 and 720 future steps. This experiment aims to investigate how different lengths of historical data affect the model’s ability to generate accurate predictions for both short-term (24 steps) and long-term (720 steps) forecasts.

Figure~\ref{res:combineexp2} provides the result of the second experiment, which compares the performance of the proposed GLinear with other state-of-the-art linear predictors (NLinear~\cite{zeng2023transformers}, DLinear~\cite{zeng2023transformers}, and RLinear~\cite{li2023revisiting}) under different input prediction lengths to understand how much historical data is enough for short- and long-term forecasting.  MSE results were computed for two prediction lengths (24 and 720 time steps) to analyze short-term and long-term forecasting performance, respectively.

It is worth noting that, similar to the previous experiment, the proposed GLinear model outperforms all other linear predictors for the Electricity and Traffic datasets. For the Weather dataset, both GLinear and RLinear show nearly identical forecasting performance. 

The performance comparison of predictors on the ETTh1 dataset reveals interesting insights about the proposed GLinear model. Specifically, GLinear's performance declines as it uses more historical data. This behavior is expected, as GLinear processes each channel independently with a compact two-layer MLP, which excels at modeling local temporal patterns but has limited capacity to capture very long-term dependencies. Across the four datasets and horizons evaluated, \emph{GLinear} achieves competitive performance and often outperforms NLinear, DLinear, RLinear, and the representative Transformer baseline (Autoformer), while exhibiting efficient training and inference.

%% file: sections/conclusion.tex
In this paper, we have introduced GLinear, a novel and data-efficient architecture for multivariate time series forecasting (TSF) that leverages periodic patterns to enhance prediction accuracy while requiring less historical data compared to existing linear models. Our experiments across four datasets (ETTh1, Electricity, Traffic, and Weather) demonstrate that GLinear not only achieves competitive performance but also outperforms state-of-the-art models such as NLinear, DLinear, and RLinear, as well as Transformer-based models like Autoformer, in various TSF scenarios.

Overall, GLinear represents a promising step towards simpler, more efficient architectures for TSF tasks, achieving robust results with lower computational and data requirements. We believe that this approach opens new avenues for the development of efficient models for time series analysis, offering both better accuracy and computational savings. 

Future work could explore enhancements such as dilated convolutions, lightweight attention modules, or temporal decomposition to improve long-term forecasting performance, and apply the GLinear architecture to other time series-related tasks, such as anomaly detection or forecasting in different domains.
Additionally, the periodic pattern extraction mechanism can be integrated into other deep learning models to enhance their efficiency and predictive performance.

%% file: bio.tex
\newpage
\section{Biography Section}
\vspace{-10em}
\begin{IEEEbiography}[{\includegraphics[width=1in,height=1.25in,clip,keepaspectratio]{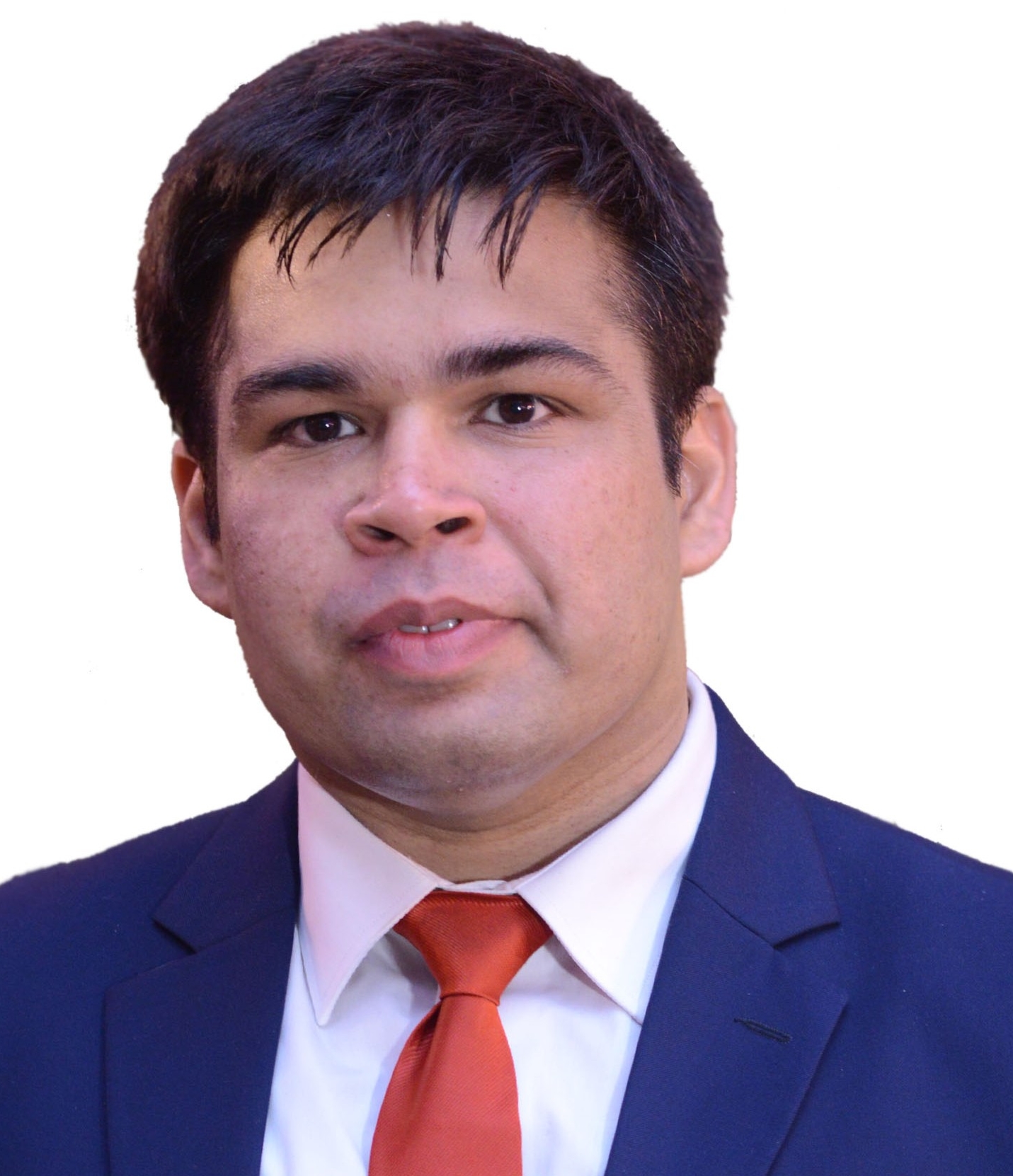}}]
{\textbf{SYED TAHIR HUSSAIN RIZVI}} received his Ph.D. in Computer and Control Engineering from Politecnico di Torino, Italy, in 2018. He has worked as a Postdoctoral Researcher at the Department of Electronics and Telecommunications (DET), Politecnico di Torino, and the Department of Electrical Engineering and Computer Science, University of Stavanger (UiS), Norway. His research interests include the efficient implementation of algorithms on embedded systems and applications of machine learning.  \vspace{-5em}
\end{IEEEbiography}
\vspace{-5em}
\begin{IEEEbiography}[{\includegraphics[width=1in,height=1.25in,clip,keepaspectratio]{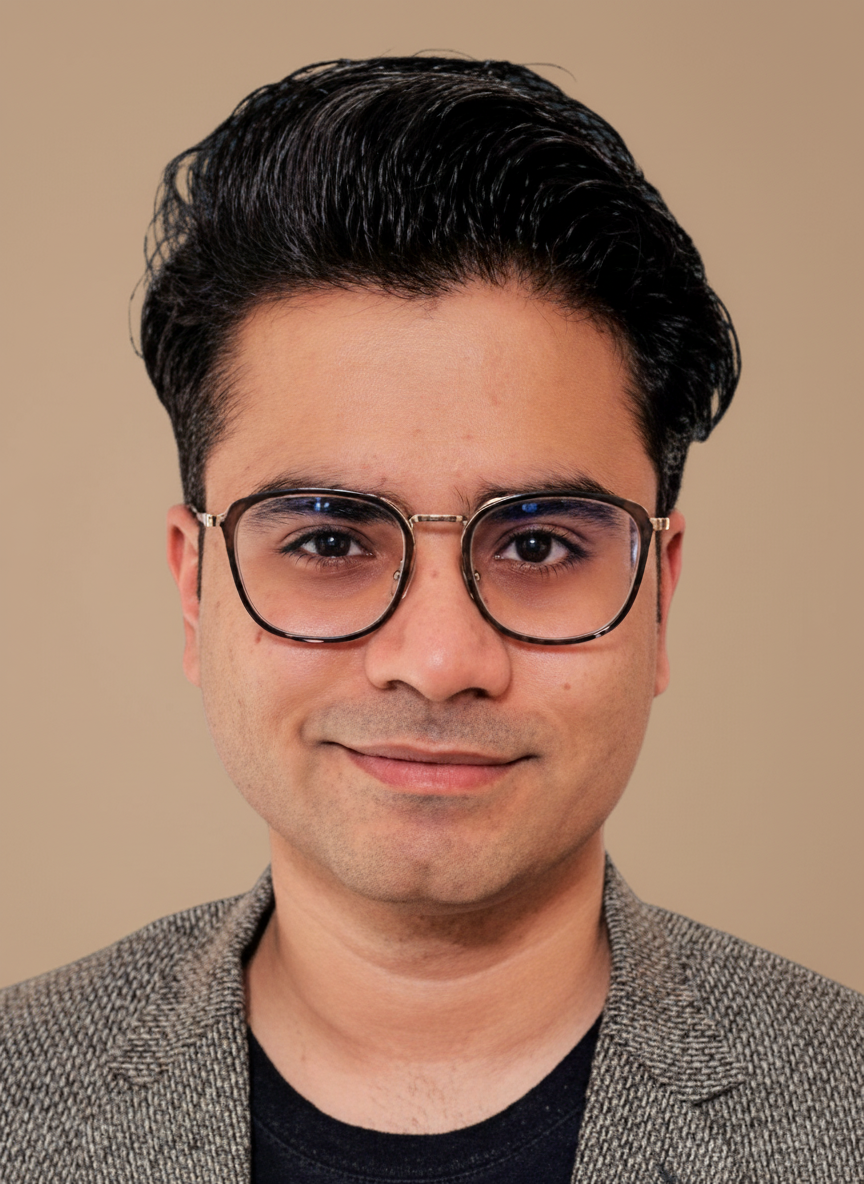}}]
{\textbf{NEEL KANWAL}} 
received his bachelor's degree in Electronics Engineering from the National University of Computer \& Emerging Sciences in 2015 and M.Sc. in Communication \& Computer Networks Engineering from Politecnico di Torino in 2020. He got his Ph.D. at the University of Stavanger (UiS), Norway, in AI for medical image analysis in 2024. He is a postdoc researcher of the Biomedical Data Analysis (BMD) Laboratory at the Department of Electrical Engineering and Computer Science, UiS. Previously, he worked as a laboratory engineer at FAST University, Karachi. The doctoral research was part of CLARIFY (European Marie-Curie Program), which aimed to develop a robust diagnostic environment for digital pathology. He is currently a postdoctoral researcher at the UiS, working on fetal heart rate time series analysis. \vspace{-5em}
\end{IEEEbiography}
\vspace{-5em}
\begin{IEEEbiography}[{\includegraphics[width=1in,height=1.25in,clip,keepaspectratio]{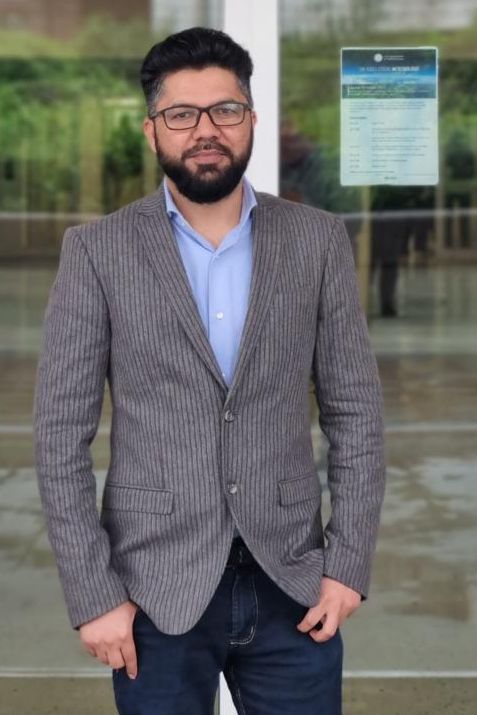}}]
{\textbf{MUDASSIR NAEEM}} 

received his bachelor's degree in Electronics from the Quaid-i-Azam University in 2010 and Master's degree in Telecommunication \& Networks from IQRA University in 2015. He got his Ph.D. at the University of Naples Parthenope, Italy, in Information and Communication Technology and Engineering in 2021. He is the coordinator of the Artificial Intelligence and Robotics (AIR) Laboratory at the Università Giustino Fortunato. Previously, he worked as a Researcher at ICAR-CNR, Naples.  He is currently a Lecturer and Researcher at Università Giustino Fortunato. 
\end{IEEEbiography}